\documentclass{amia}
\usepackage{lipsum} %Remove if not needed
\setcitestyle{maxnames=1,minnames=1}

\begin{document}

\title{Temporally Phenotyping GLP-1RA Case Reports with Large Language Models: A Textual Time Series Corpus and Risk Modeling}

\author{Sayantan Kumar, PhD,$^1$, Jeremy C. Weiss, MD,PhD$^1$ }

\institutes{
    $^1$ National Library of Medicine, Bethesda, Maryland, USA
}

\maketitle

\section*{Abstract}
\vspace{-2.5mm}
Type 2 diabetes case reports describe complex clinical courses, but their timelines are often expressed in language that is difficult to reuse in longitudinal modeling. To address this gap, we developed a textual time-series corpus of 136 PubMed Open Access single-patient case reports involving glucagon-like peptide 1 receptor agonists, with clinical events associated with their most probable reference times. We evaluated automated LLM timeline extraction against gold-standard timelines annotated by clinical domain experts, assessing how well systems recovered clinical events and their timings. The best-performing LLM produced high event coverage (GPT5; 0.871) and reliable temporal sequencing across symptoms (GPT5; 0.843), diagnoses, treatments, laboratory tests, and outcomes. As a downstream demonstration, time-to-event analyses in diabetes suggested lower risk of respiratory sequelae among GLP-1 users versus non-users (HR=0.259, p<0.05), consistent with prior reports of improved respiratory outcomes. Temporal annotations and code will be released upon acceptance.

%%%%%%%%%%%%%%%%%%%%%%%%%%%%%%%%%%%%%%%%%%%%%%%%%%%%%%%%%%%
%%%%%%%%%%%%%%%%%%%%%%%%%%%%%%%%%%%%%%%%%%%%%%%%%%%%%%%%%%%
\section*{Introduction}
\vspace{-3mm}
Type 2 diabetes (T2D) is a chronic metabolic disorder characterized by insulin resistance, progressive $\beta$-cell dysfunction, and hyperglycemia, which together increase the risk of complications including cardiovascular disease, nephropathy, and neuropathy \cite{defronzo2015combination}. Glucagon-like peptide-1 receptor agonists (GLP-1RAs) like liraglutide and semaglutide are central to current treatment because they enhance insulin secretion, suppress glucagon release, slow gastric emptying, and promote weight loss. \cite{marso2016semaglutide,wilding2021once}. However, their long-term effects on diabetes progression remain unclear. Because most studies emphasize diagnosis and short-term outcomes, it is important to reconstruct longitudinal, temporally aligned trajectories in GLP-1RA case reports to support long-term risk forecasting and to capture heterogeneity in treatment response and downstream complication risk.

Traditional studies of T2D progression and GLP-1RA–associated outcomes largely rely on structured electronic health records (EHRs), claims data, and trial cohorts, which provide timestamped measurements but often lack an explicit representation of medication-centered disease dynamics over time. 
%Although EHR resources such as MIMIC-III and MIMIC-IV provide structured variables and clinical notes \cite{johnson2016mimic, johnson2023mimic}, they capture only a subset of clinically meaningful events (e.g., vitals and laboratory results) and require sophisticated natural language processing to recover key temporal details from unstructured text, including treatment context, adherence, and adverse-event narratives \cite{viani2021temporal, cheng2023typed}. 
Although EHR resources such as MIMIC-III and MIMIC-IV provide both structured variables and clinical notes \cite{johnson2016mimic, johnson2023mimic}, they are largely derived from critical-care populations and emphasize short-term inpatient trajectories, which limits their ability to capture outpatient GLP-1RA treatment courses and long-horizon outcomes.
As a result, even when timestamps exist, reconstructing GLP-1RA initiation, dose changes, tolerability, and downstream complication onset as coherent trajectories remains challenging.

By contrast, unstructured clinical narratives such as published case reports often contain detailed descriptions of GLP-1RA treatment courses and clinical evolution, but event timing is typically expressed in relative terms within free text (e.g., “on day 3 of hospitalization,” “two weeks after starting semaglutide”). Progress in clinical temporal reasoning has been limited by the scarcity of large, richly annotated corpora. Early efforts—including the 2012 i2b2 challenge and Clinical TempEval—advanced temporal relation extraction from clinical notes \cite{sun2013evaluating,gumiel2021temporal} but were constrained by small datasets (e.g., 310 documents) drawn from single institutions, affecting performance and generalizability. Moreover, approaches that rely mainly on metadata timestamps (e.g., admission or discharge dates) miss the fine-grained event sequences described in text—such as medication indications, side effects, and clinical decision points—which are critical for understanding treatment response and downstream risk in real-world GLP-1RA use. Together, these limitations motivate building temporally annotated case-report corpora to align narrative events with medication exposure and outcomes, enabling medication-anchored trajectories for risk prediction and personalized treatment planning when structured data are incomplete.

\begin{figure}[!htbp] % 'r' for right, 'l' for left
    \centering
    \includegraphics[width=\linewidth]{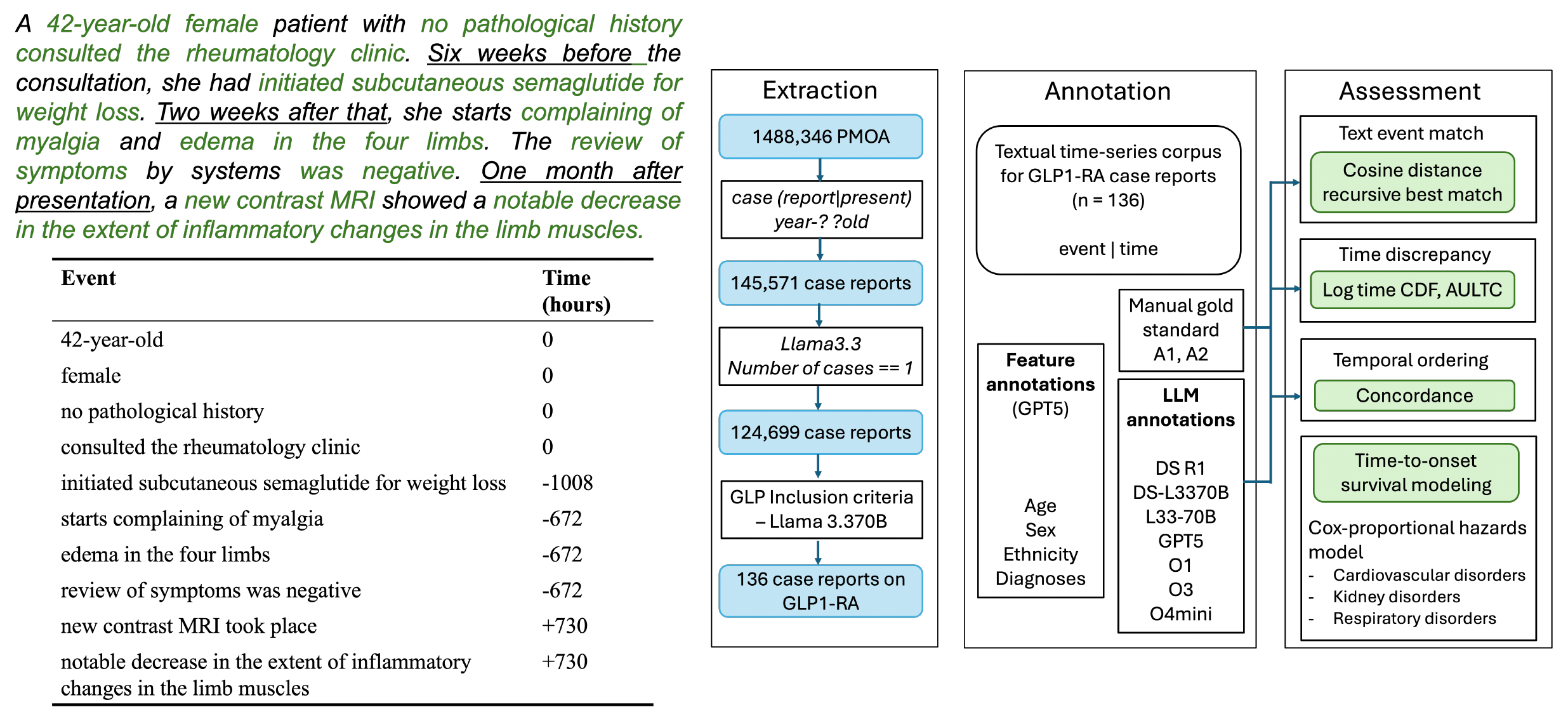} % Adjust width
    \caption{\textbf{Left:} Example case report (top) with text-ordered event-time tuples (bottom). Clinical events and temporal cues are marked in green and underline respectively. \textbf{Right:} Overview of our pipeline. \underline{Left panel:} filtering the PMOA corpus to identify case reports of patients administered GLP1-RA medications. \underline{Middle panel:}textual time series generation for each case report via LLM prompting and the creation of structured feature subsets. \underline{Right panel:} Evaluation process, including text event matching, time discrepancy analysis (log- time CDF, AULTC), temporal order concordance and time-to-onset survival modeling of selected outcomes.
    }
    \label{fig:fig1_worflow_with_example}
    \vspace{-2mm}
\end{figure}
%-------------------------------------------
\underline{Contributions}

Our work presents the following contributions. \textbf{First}, we introduce a novel GLP-1RA textual time-series corpus derived from PubMed Open Access (PMOA) case reports, enabling fine-grained modeling of diabetes progression and GLP-1RA treatment response directly from unstructured narratives. We convert each case report into a structured, time-stamped clinical timeline using an \emph{LLM-as-annotator} approach. We then benchmark multiple LLMs to evaluate both their ability to extract clinically meaningful findings and their ability to accurately associate those findings with relative timestamps. \textbf{Second}, to support rigorous evaluation and reuse, we curate a manual gold standard in which two clinically-trained domain experts extract reference timelines and assess whether GLP-1RA is the primary intervention. \textbf{Third}, we characterize the cohort’s demographic and clinical breadth, including medication patterns and frequently observed diagnoses and comorbidities. \textbf{Fourth}, we demonstrate downstream clinical utility through time-to-onset survival modeling, using a Cox proportional hazards model to examine associations between GLP-1RA exposure status and time-to-onset of kidney, cardiovascular, and respiratory outcomes in T2D case reports. \textbf{Finally}, we release both the LLM-extracted timelines and expert annotations as a pilot benchmark corpus to support future research on temporal extraction and longitudinal clinical modeling.

%%%%%%%%%%%%%%%%%%%%%%%%%%%%%%%%%%%%%%%%%%%%%%%%%%%%%%%%%%%
\section*{Related Work}
\vspace{-3mm}
Large structured real-world data sources such as de-identified EHR databases, administrative claims, and trial cohorts are widely used to study T2D progression and GLP-1RA–associated outcomes \cite{brito2025glp}. These resources provide explicit timestamps for encounters, prescriptions, and laboratory measurements, but they often capture only a subset of clinically meaningful events and omit the narrative context needed to reconstruct medication-centered trajectories—such as indications for tolerability, adherence, and adverse-event descriptions \cite{young2023treatment,foer2023association}. Moreover, timing is central for downstream time-to-onset analyses: imprecise alignment of treatment exposure and outcome onset can obscure clinically meaningful patterns. These limitations motivate complementary text-derived, time-resolved representations that leverage the detailed event sequences described in clinical narratives when structured event streams are incomplete or unavailable. 

% Several works have used the i2b2 (a competition subset of MIMIC-II/III) to construct timelines of clinical concepts \cite{leeuwenberg2020towards,frattallone2024using}. In contrast, we assign timestamps to \emph{clinical findings} rather than i2b2 concept spans \cite{uzuner20112010}, which enables more specific, context-preserving events in settings where additional structured context is absent. Unlike \cite{frattallone2024using}, we operate in a text-only setting, which is necessary for the Pubmed Open-Access Subset where no complementary structured data streams are available.
Extracting timelines from clinical narratives is a challenging biomedical NLP task. The i2b2 2012 challenge introduced annotated datasets for temporal relation extraction from discharge summaries \citep{uzuner20112010}. Later work linked clinical events to timestamps or temporal expressions \citep{leeuwenberg2020towards, frattallone2024using}, often assuming pre-defined event spans. We assign timepoints directly to findings in full-length case reports, enabling finer temporal resolution. In contrast to methods relying on structured EHR data, we operate purely on text—essential for sources like PubMed that lack structured metadata. By directly supervising event–time alignment, we mitigate limitations of span-based annotations.

Broadly in the medical domain, LLMs are showing promise in tasks such as summarization and some evidence suggests that medical fine-tuning may not consistently improve over corresponding foundation models \cite{jeong-etal-2024-medical}. Motivated by these findings, we primarily evaluate foundation LLMs and examine how scale and instruction-tuning affect temporal reconstruction from clinical narratives, where timing is often implicit and depends on long-range context. This makes it feasible to extract time-resolved event sequences directly from text and support downstream longitudinal modeling when structured timestamps are incomplete or unavailable.

%%%%%%%%%%%%%%%%%%%%%%%%%%%%%%%%%%%%%%%%%%%%%%%%%%%%%%%%%%%
%%%%%%%%%%%%%%%%%%%%%%%%%%%%%%%%%%%%%%%%%%%%%%%%%%%%%%%%%%%
\section*{Methods}
\vspace{-3mm}
Our pipeline comprises the following key components: dataset extraction, textual time series annotation, and comprehensive evaluation of the annotations (Figure~\ref{fig:fig1_worflow_with_example} right). 
\vspace{-2mm}

\subsection*{Data extraction}
\vspace{-2mm}
%\vspace{-3pt}
\begin{sloppypar}
We use the PubMed Open Access (PMOA) repository (1,488,346 manuscripts), using plain text files from the December 17, 2024 release in \texttt{oa\_noncomm}. Following prior work \cite{wang2025large, noroozizadeh2025forecasting,noroozizadeh2025pmoa}, we extract body text between ``\texttt{==== Body}'' and ``\texttt{==== Ref}'' and identify candidate case reports via the case-insensitive regular expressions \textcolor{teal}{\texttt{case (report|present)}} and \textcolor{teal}{\texttt{year-? ?old}}, yielding 145,571 candidates.
%We then restrict to single-patient case reports using an LLM-based filter (124,699 reports) and identify a GLP-1RA cohort by prompting \texttt{Llama 3.3 70B Instruct} to retain reports that satisfies either of the inclusion criteria:(i) document GLP-1RA administration (e.g., semaglutide, liraglutide, exenatide, dulaglutide, tirzepatide), (ii) describe clinical effects, adverse reactions, or outcomes related to GLP-1RA therapy, and (iii) use GLP-1RA in diagnosis, prognosis, or therapeutic decision-making. This yields 136 GLP-1RA case reports for downstream timeline extraction and analysis (Figure~\ref{fig:fig1_worflow_with_example} right).
 We then restrict to single-patient case reports using an LLM-based filter (124,699 reports) and identify a GLP-1RA cohort by performing keyword-based matching against the full text using a curated lexicon. The lexicon included (i) class-level expressions matched via regular expressions (e.g., \texttt{“glp1”, “glp 1 receptor agonist”, “glp 1 analog”, “glp 1 ra”, “glucagon like peptide 1”}) and (ii) explicit drug-name matches (\texttt{semaglutide, liraglutide, exenatide, dulaglutide, lixisenatide, albiglutide, efpeglenatide, tirzepatide}). 136 case reports containing any lexicon term were retained as GLP-1RA candidate case reports for downstream timeline extraction and analysis (Figure~\ref{fig:fig1_worflow_with_example} right).
\end{sloppypar}
%\vspace{-3pt}

\vspace{-2mm}
\subsection*{Textual time-series annotations}
\vspace{-2mm}
\begin{sloppypar}
We used large language models (LLMs; \texttt{DeepSeek R1, DeepSeek-R1-Distill-Llama-70B, Llama3.3-70B-Instruct, GPT5, O1, O3, O4mini}) to extract \emph{textual time series (TTS)} from 136 GLP PMOA case reports. Given a clinical text document $T$, we define TTS as $S=\{(e_1,t_1),(e_2,t_2),\ldots,(e_n,t_n)\}$, where $e_i$ is a clinical finding (a contiguous text span describing a temporally localizable patient event) and $t_i \in \mathbb{R}$ is the event time in hours relative to a case-specific reference point ($t=0$). We define the reference point as hospital admission when explicitly described; otherwise we use the earliest documented clinical encounter or presentation in the report. Events before the reference point receive negative timestamps and events after receive positive timestamps. 
\end{sloppypar}
In this work, a clinical event is any patient-specific, health-related mention that is semantically self-contained and relevant to the clinical course, including symptoms/signs, diagnoses, procedures and diagnostic tests, treatments/medication administrations, major clinical states, and outcomes (Figure \ref{fig:fig1_worflow_with_example} left). We also include explicitly stated pertinent negatives and termination events (e.g., no shortness of breath,'' antibiotics discontinued''). Demographic attributes mentioned in the narrative (e.g., age and sex) are recorded as events at the reference time ($t{=}0$). We exclude non–patient-specific contextual text, such as general background statements and literature discussion. 

Following our prior work \cite{wang2025large}, we keep findings as minimally edited, context-rich spans rather than restricting them to short i2b2-style concept phrases \cite{sun2013evaluating,uzuner20112010}. In particular, (1) findings may extend beyond single prepositional phrases for specificity (e.g., ``pain in chest that radiates substernally''), and (2) conjunctive mentions are split into separate findings when this improves clarity (e.g., ``metastases in liver and pancreas'' $\rightarrow$ \{``metastasis in liver'', ``metastasis in pancreas''\}). Natural-language time expressions are normalized to hour offsets using start times (e.g., ``three-day history of fever'' $\rightarrow t=-72$); coarse expressions such as ``hospital day 2'' are mapped using 24-hour increments, and vague phrases are assigned approximate offsets consistent with narrative order and surrounding cues.

\vspace{-2mm}
\subsection*{Structured feature subsets: demographics and diagnoses}
\vspace{-2mm}

For demographics, we used GPT5 (best-performing LLM selected from Figure \ref{fig:tts_eval}) to extract age, sex, and ethnicity from each case report. The model was instructed to report age in years, to encode sex as Male, Female, a free-text value, or Not Specified, and to assign ethnicity using U.S. Census categories when possible (otherwise Not Specified).

To summarize diagnostic coverage, we also prompted GPT5 (as a physician) to generate a patient-specific diagnosis list for each report. The prompt emphasized final diagnostic labels (e.g., “acne,” “DRESS syndrome”) rather than individual findings (e.g., “rash,” “leukocytosis”), requested that the primary diagnosis appear first, and required one diagnosis per line with no additional explanation.
To standardize LLM-extracted diagnoses, we mapped each free-text diagnosis to the Unified Medical Language System (UMLS) using the ScispaCy entity linker (\texttt{en\_core\_sci\_lg}) with the UMLS knowledge base. For each diagnosis string, we retained only candidate mappings with confidence score greater $>0.85$; when multiple concepts were returned, we selected the highest-ranked Concept Unique Identifier (CUI) and used its canonical UMLS name for downstream analyses.

To summarize diagnosis patterns at a higher level, we additionally grouped UMLS-normalized diagnoses into the following coarse disease categories ( \href{https://www.heart.org/en/about-us/heart-and-stroke-association-statistics}{cardiovascular diseases}, \href{https://www.cdc.gov/oral-health/about/gum-periodontal-disease.html}{oral \& dental health}, \href{https://pubmed.ncbi.nlm.nih.gov/35794458/}{respiratory diseases}, \href{https://www.nimh.nih.gov/health/statistics/mental-illness}{dementia \& mental health}, \href{https://www.niddk.nih.gov/health-information/health-statistics/digestive-diseases}{digestive \& hepatic disorders}, \href{https://www.cdc.gov/nchs/products/databriefs/db497.htm}{joint \& bone disorders}, \href{https://www.cdc.gov/kidney-disease/php/data-research/index.html}{kidney disease}, \href{https://www.cdc.gov/nchs/products/databriefs/db516.htm}{diabetes}, \href{https://www.cdc.gov/nchs/products/databriefs/db519.htm}{anemia}, \href{https://www.cdc.gov/nchs/fastats/cancer.htm}{cancer}, \href{https://www.cdc.gov/sepsis/about/index.html}{sepsis}) and computed category prevalence as the fraction of case reports containing at least one diagnosis mapped to each category. These comparisons are intended to contextualize frequencies across categories rather than estimate population incidence. 

\vspace{-2mm}
\subsection*{Manual gold standard annotations}
\vspace{-2mm}

We construct a manual gold-standard reference set of 136 GLP-1RA case reports, annotated independently by two clinically trained expert annotators using detailed annotation guidelines summarized as follows. Annotators select exact text spans for events (no paraphrasing), with two controlled exceptions: (i) conjunctive mentions can be split into clinically meaningful individual events, and (ii) a brief prefix such as “history of” can be added when necessary to make an event interpretable in isolation. Annotators also ensure each clinical event is semantically self-contained (e.g., using “SCC of the lung” rather than “SCC” to avoid ambiguity with other entities such as cutaneous squamous cell carcinoma or sickle cell crisis), avoid duplicate event mentions across sections, and include explicitly stated pertinent negatives and termination events (e.g., “no shortness of breath” or “denies chest pain”). Each event is assigned a relative timestamp anchored to a standardized case reference point (admission when stated; otherwise the earliest unambiguous clinical encounter), recording interval expressions at their start time; timestamps are recorded in hours.

\vspace{-2mm}
\subsection*{Comparing LLM and manual timelines - event matching and temporal performance}
\vspace{-2mm}

We assess LLM-extracted timelines by comparison to the manual gold standard at both the event and temporal levels. First, we align predicted and reference events \emph{within each case report} using a recursive best-match procedure adapted from prior work \cite{wang2025large}, which greedily selects the closest unmatched predicted--reference pair under a text-similarity metric, retains the pair if it passes a similarity threshold, and removes both events before continuing. This yields an efficient one-to-one matching between timelines of differing lengths. Using this alignment, we report \textbf{event match rate}, defined as the fraction of reference events that have a matched prediction.

For semantic matching, we compared multiple similarity measures and found cosine similarity of PubMedBERT sentence embeddings to perform best; we therefore treat a predicted event as a match when its cosine distance to a reference event is $\le 0.1$. Among matched events, we evaluate temporal quality in two complementary ways. \textbf{Concordance (c-index)} measures ordering agreement: the probability that pairs of matched events appear in the same temporal order in the prediction as in the reference timeline. \textbf{Timestamp discrepancy} measures absolute differences between predicted and reference times for matched events. Because these errors can span orders of magnitude, we summarize them on a log scale and report the \textbf{Area Under the Log-Time CDF (AULTC)\cite{noroozizadeh2025reconstructing}}, which quantifies how concentrated timestamp errors are near zero (higher is better).

\vspace{-2mm}
\subsection*{Time-to-onset survival modeling}
\vspace{-2mm}

\begin{sloppypar}
To demonstrate downstream clinical utility of GLP-1RA textual time series, we performed time-to-onset analyses for kidney, cardiovascular, and respiratory outcomes, using group definitions designed to examine the \emph{association} between GLP-1RA exposure and subsequent outcome onset. We defined a GLP-1RA \emph{treatment} cohort as PMOA case reports in which the patient both (i) had diabetes-related diagnoses and (ii) received a GLP-1RA medication. Diabetes status was identified by searching for diabetes-related keywords in the LLM-extracted diagnosis list: \texttt{[diabetes,diabetic,dm,type 1,type 2, insulin,metabolic syndrome, hyperglycem, hypoglycem, ketoacidosis, hba1c]}. This yielded 120 GLP-1RA case reports. Time was measured from the case reference time ($t{=}0$) and converted to months for survival analyses. To reduce immortal time bias due to delayed treatment initiation, we required the first GLP-1RA administration to occur within 72 hours of $t{=}0$, yielding 82 treated cases; the remaining 38 cases with GLP-1RA initiation $>72$ hours after $t{=}0$ were excluded from the treatment cohort.

We defined a \emph{comparison} cohort consisting of (i) diabetes case reports with no GLP-1RA exposure and (ii) the 38 cases with GLP-1RA initiation $>72$ hours after $t{=}0$ (treated as unexposed at baseline). Additional comparison cases were sampled at random from the broader PMOA single-patient pool to achieve a 5:1 control-to-treatment ratio. Outcomes were identified using keyword-based definitions (Figure \ref{fig:survival} right) applied to the time-ordered TTS events; LLM-derived diagnosis-list matching was explored initially but was not used in the final survival analyses due to lower recall. 
%For each case, event time was defined as the earliest timestamp of any matching outcome event in the extracted timeline; cases without a detected outcome were right-censored at the end of follow-up, defined as the maximum timestamp in the time-ordered TTS.
For each case, event time was defined as the earliest timestamp of any matching outcome event in the extracted timeline. Because case reports often document pre-existing conditions at presentation, outcome events occurring at $t{=}0$ should be interpreted as baseline documentation (prevalent at reference time) rather than true incident onset; thus, our endpoint is best viewed as time to first appearance in the extracted timeline. Cases without a detected outcome were right-censored at the end of follow-up, defined as the maximum timestamp in the time-ordered TTS.
\end{sloppypar}

We estimated unadjusted time-to-onset curves using the Kaplan--Meier estimator and fit Cox proportional hazards models with a baseline binary indicator of GLP-1RA exposure, adjusting for age and sex as covariates (sex encoded via one-hot indicators). All survival analyses were implemented in Python using the \texttt{lifelines} package \cite{davidsonpilon2019lifelines}. Adjusted survival curves were generated from the fitted Cox model by predicting survival for a representative individual with mean age and modal sex under treatment versus comparison exposure, and uncertainty bands were computed via nonparametric bootstrap resampling (percentile 2.5/97.5\%; 500 resamples; resamples with convergence failures were omitted). Throughout, we interpret hazard ratios as associative rather than causal, given the observational nature of case reports and limited covariate information.

%%%%%%%%%%%%%%%%%%%%%%%%%%%%%%%%%%%%%%%%%%%%%%%%%%%%%%%%%%%

\begin{figure}[t] % 'r' for right, 'l' for left
    \centering
    \includegraphics[width=\linewidth]{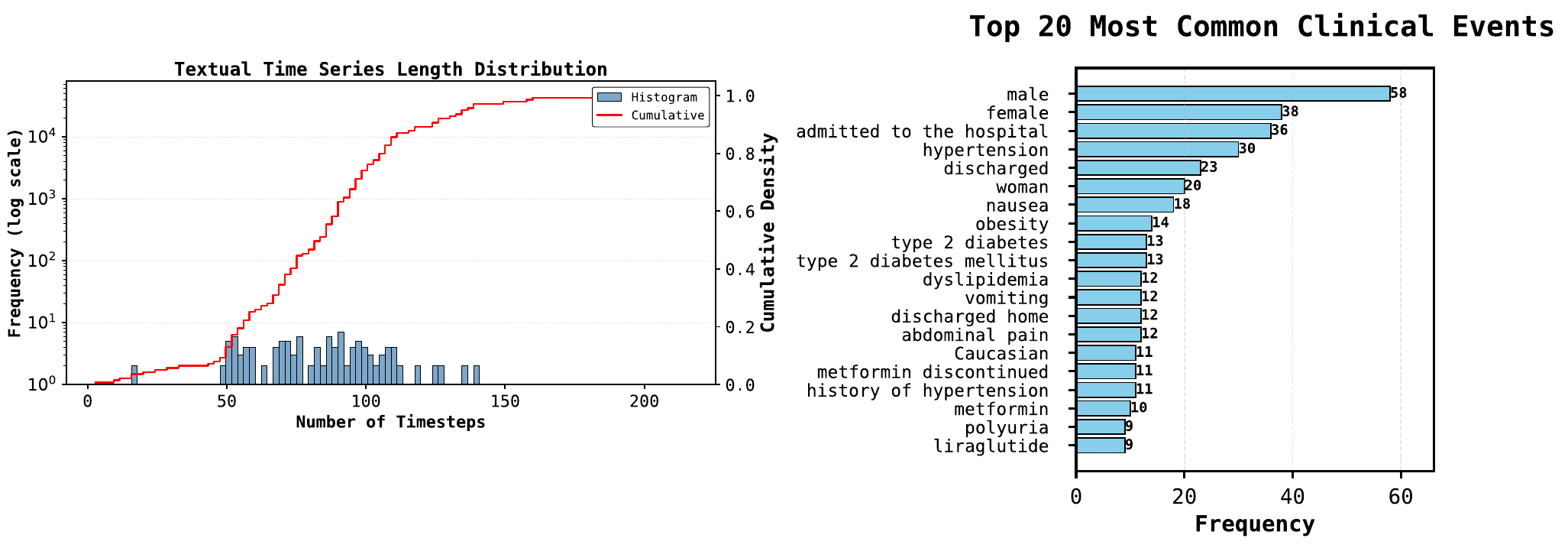} % Adjust width
    \caption{\textbf{a} Distribution of time series lengths
    (timesteps) across the dataset. \textbf{b} Most frequently occurring events across all case reports.
    }
    \label{fig:tts_stats}
    \vspace{-2mm}
\end{figure}

%%%%%%%%%%%%%%%%%%%%%%%%%%%%%%%%%%%%%%%%%%%%%%%%%%%%%%%%%%%
\section*{Results}

\vspace{-2mm}
\subsection*{Descriptive statistics: GLP TTS cohort}
\vspace{-2mm}

\paragraph{Demographics:} Across the PMOA GLP cohort, the median age is 49 years (IQR $\pm$ 17 years). The inferred sex distribution is nearly balanced (49\% male, 49\% female), with 2\% not specified. Ethnicity is infrequently reported: 78\% of case reports do not document ethnicity, and among those that do, White/Caucasian (7\%) is most common, followed by Asian (6\%) and White (6\%). The relatively higher proportion of Asian patients likely reflects the international authorship of PubMed case reports rather than representing any single healthcare system. Taken together, these demographics suggest a cohort largely reflecting mid-life adults, consistent with the typical age range in which cardiometabolic comorbidities and GLP-1RA use are frequently discussed in clinical narratives.

\paragraph{Clinical events:} The GLP-1RA textual time series (TTS) extracted from PMOA case reports vary widely in length (Figure~\ref{fig:tts_stats}a). Most timelines contain on the order of tens to low hundreds of timesteps, with the highest density concentrated roughly between 50 and 110 events and a right-skewed tail extending beyond 200 timesteps. This variability likely reflects differences in reporting detail and the number of clinically relevant findings explicitly documented as time-localized events (e.g., detailed hospital courses vs. brief presentations).

\begin{sloppypar}
Figure~\ref{fig:tts_stats}b shows the 20 most frequent extracted event strings. The most common events include demographic descriptors (e.g., \texttt{male}, \texttt{female}, \texttt{woman}) and care milestones (e.g., \texttt{admitted to the hospital}, \texttt{discharged}, \texttt{discharged home}), as well as prevalent cardiometabolic comorbidities and diabetes-related treatments consistent with a GLP-1RA cohort (e.g., \texttt{hypertension}, \texttt{obesity}, \texttt{type 2 diabetes}, \texttt{type 2 diabetes mellitus}, \texttt{dyslipidemia}, \texttt{metformin}, \texttt{metformin discontinued}, \texttt{liraglutide}). Gastrointestinal symptoms (\texttt{nausea}, \texttt{vomiting}, \texttt{abdominal pain}) also appear among frequent events, reflecting commonly documented symptomatology in these narratives. These frequencies are computed from extracted \emph{event strings} and therefore reflect only information explicitly stated and captured by the pipeline; accordingly, demographic events such as \texttt{male} and \texttt{female} need not sum to the number of case reports.
\end{sloppypar}

\paragraph{Temporal statistics:} Temporal coverage also varies substantially across reports. The mean case duration is 3{,}930 days (11 years) and the median is 2{,}565 days (7 years); the middle 50\% of cases span 438 to 5{,}997 days (16 years), and the 99th percentile reaches 7{,}380 days (20 years). This long follow-up is consistent with case reports that often summarize multi-year disease histories and longitudinal treatment courses rather than short, single-encounter episodes. Events frequently co-occur at the same recorded times: on average, 71.4\% of events share a timestamp with at least one other event, and the median timestamp uniqueness ratio is 0.16, consistent with documentation of multiple findings at a single timepoint (e.g., grouped laboratory results or bundled assessments). In addition, approximately 24\% of events have negative timestamps, reflecting pre-admission history such as prior diagnoses or treatments. Together, these statistics characterize the density and temporal heterogeneity of patient trajectories in the GLP PMOA cohort.

%%%%%%%%%%%%%%%%%%%%%%%%%%%%%%%%%
\begin{figure}[t] % 'r' for right, 'l' for left
    \centering
    \includegraphics[width=\linewidth]{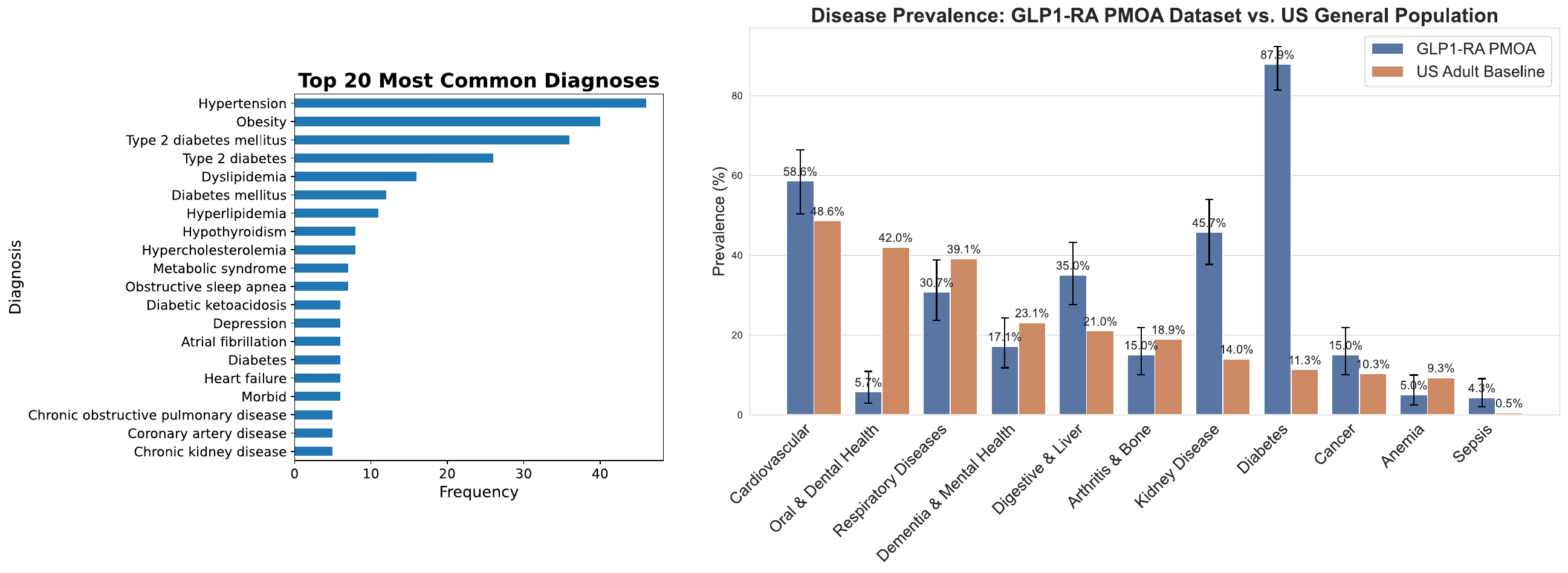} % Adjust width
    \caption{Frequency and prevalence patterns of UMLS-normalized diagnoses in PMOA-TTS. (\textbf{a}) Top 20 diagnoses by frequency, reported using canonical UMLS names. (\textbf{b}) Prevalence of broad disease categories in PMOA-TTS compared with published U.S.\ adult baseline estimates, highlighting systematic differences between case-report cohorts and general-population distributions. 
    }
    \label{fig:diagnosis_stats}
    \vspace{-2mm}
\end{figure}
%%%%%%%%%%%%%%%%%%%%%%%%%%%%%%%%%

\vspace{-2mm}
\subsection*{Diagnoses distribution analyses}
\vspace{-2mm}

\begin{sloppypar}
Figure~\ref{fig:diagnosis_stats}a shows the 20 most frequent UMLS-normalized diagnoses in the GLP-1RA PMOA cohort. The list is dominated by cardiometabolic conditions—particularly Hypertension and Obesity—and multiple diabetes labels (e.g.,Type 2 diabetes, Diabetes mellitus). Other common comorbidities include lipid disorders (Dyslipidemia, Hyperlipidemia), Metabolic syndrome, Obstructive sleep apnea, and Chronic kidney disease, along with associated conditions such as Coronary artery disease, Atrial fibrillation, Chronic obstructive pulmonary disease, Diabetic ketoacidosis, Hypothyroidism, and Depression. Overall, this cardiometabolic profile is consistent with the clinical contexts in which GLP-1RAs are commonly prescribed and studied, and provides a face-validity check that the extracted diagnosis distributions reflect expected comorbidity patterns rather than spurious artifacts.

As shown in Figure~\ref{fig:diagnosis_stats}b, disease-group prevalence in the GLP-1RA PMOA cohort differs from U.S.\ adult baselines in a pattern consistent with case-report selection effects. Diabetes is strongly enriched (87.9\% vs.\ 11.3\%), and kidney disease is also substantially overrepresented (45.7\% vs.\ 14.0\%), alongside enrichment of digestive \& liver conditions (35.0\% vs.\ 21.0\%). Cardiovascular disease remains common (58.6\% vs.\ 48.6\%), while oral \& dental health (5.7\% vs.\ 42.0\%), respiratory diseases (30.7\% vs.\ 39.1\%), and dementia \& mental health (17.1\% vs.\ 23.1\%) appear lower than baseline. Sepsis, although rare in the general population, is enriched in the cohort (4.3\% vs.\ 0.5\%), likely reflecting the tendency of published case reports to overrepresent severe, hospitalized, or diagnostically complex presentations. The strong enrichment of diabetes (87.9\%) indicates that GLP-PMOA is largely diabetes-centered, which is expected given that GLP-1RA use in case reports is frequently discussed in the context of glycemic management. These comparisons are intended for descriptive cohort characterization rather than epidemiologic inference, since the cohort comprises published case reports and prevalence depends on diagnosis-to-group mapping choices.
\end{sloppypar}

%%%%%%%%%%%%%%%%%%%%%%%%%%%%%%%%%%%%%%%%%%%%%%%%%%%%%%%%%%%
\begin{figure}[t] % 'r' for right, 'l' for left
    \centering
    \includegraphics[width=\linewidth]{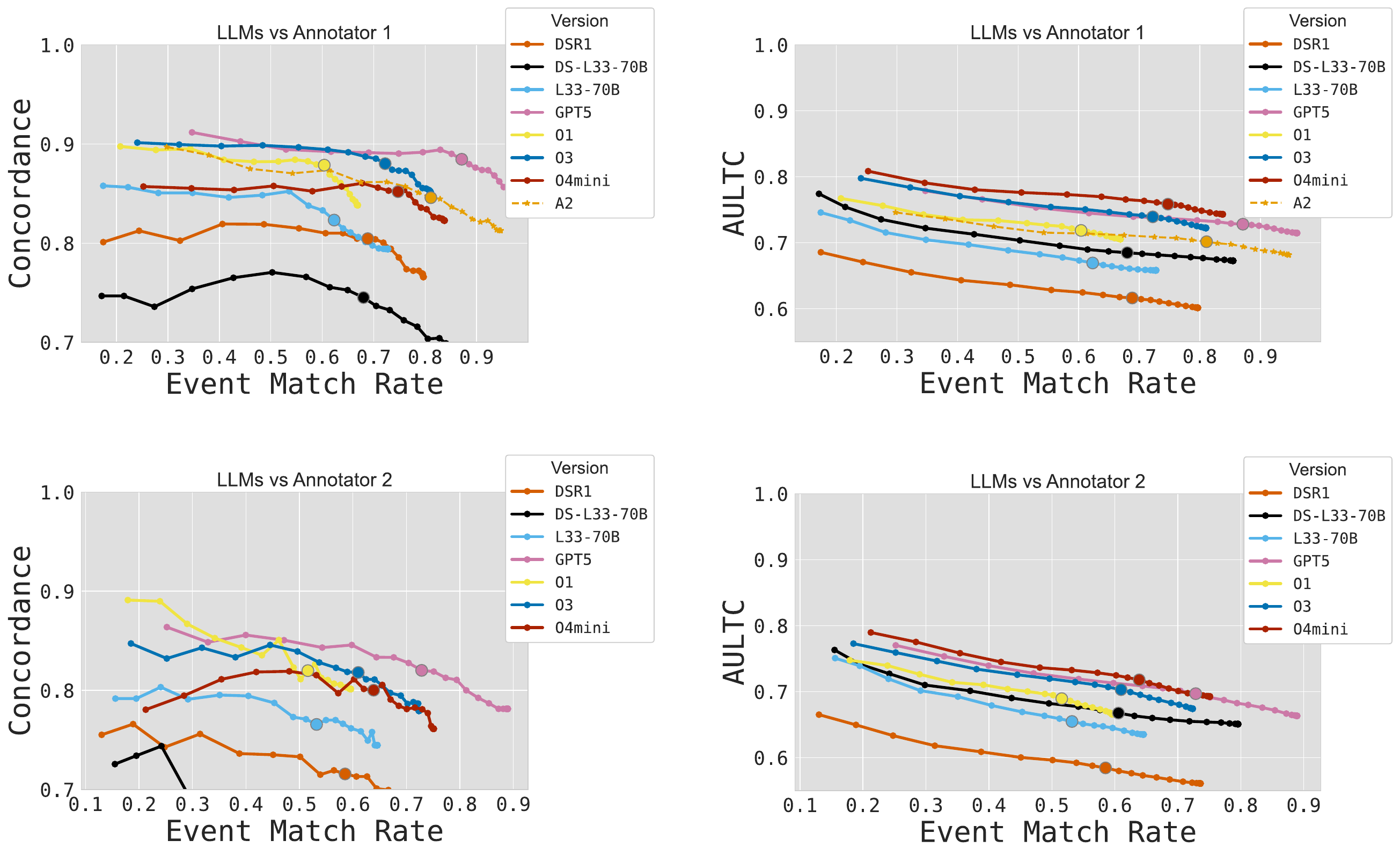} % Adjust width
    \caption{Sensitivity analysis of clinical textual time series (TTS) quality across event-matching thresholds. Performance is summarized as \textbf{concordance} (ordering agreement) and \textbf{AULTC} (timestamp accuracy) plotted against \textbf{event match rate} for comparisons to Annotator~1 (top row) and Annotator~2 (bottom row). Solid circle ($\bullet$) represents threshold of 0.1, with ticks (|) indicating 0.01 increments of the threshold in [0.01, 0.50]. Curves closer to the upper-right indicate models that recover a larger fraction of expert events while preserving temporal order and producing smaller timestamp discrepancies.
    }
    \label{fig:tts_eval}
    \vspace{-2mm}
\end{figure}
%%%%%%%%%%%%%%%%%%%%%%%%%%%%%%%%%%%%%%%%%%%%%%%%%%%%%%%%%%%

\vspace{-2mm}
\subsection*{Evaluation of Clinical Textual Time Series Quality}
\vspace{-2mm}

We performed a sensitivity analysis by varying the PubMedBERT cosine-distance threshold used for event matching, which traces a tradeoff between \textbf{event match rate} (coverage) and temporal fidelity. %Figure~\ref{fig:tts_eval} reports \textbf{concordance} and \textbf{AULTC} as functions of \textbf{event match rate} for each LLM, comparing model timelines to Annotator~1 (top row) and Annotator~2 (bottom row). 
%Curves closer to the upper-right indicate models that recover a larger fraction of expert events while preserving temporal order and producing smaller timestamp discrepancies.
Across both annotators, \texttt{GPT5} consistently provides the strongest overall tradeoff, achieving the highest match rates while maintaining high concordance and AULTC. \texttt{O3} (blue) is the most competitive alternative, exhibiting high ordering agreement over a broad match-rate range and strong timestamp accuracy, while \texttt{O4mini} attains among the best AULTC values at comparable match rates, with slightly lower concordance than the top curves. In contrast, the open-weight models (\texttt{L33-70B}, \texttt{DS-L33-70B}, \texttt{DSR1}) generally operate at lower match rates and/or show lower temporal fidelity, with \texttt{DS-L33-70B} performing worst among the evaluated models. Finally, absolute performance is consistently lower when using Annotator~2 as reference, but the relative ordering of model tradeoffs remains similar, suggesting that the observed differences are robust to the choice of expert baseline.

\paragraph{Inter-annotator agreement:}Using the same matching and temporal metrics, we also computed inter-annotator agreement (A2 vs A1) and observed 0.811 event match, 0.798 concordance (C-index), and 0.702 AULTC, providing a human baseline for expected agreement on event content and timing. Notably, the best-performing model (\texttt{GPT5}) have better event matching and temporal ordering compared to Annotator 2, indicated by higher concordance and AULTC at similar match rates.

%%%%%%%%%%%%%%%%%%%%%%%%%%%%%%%%%%%%%%%%%%%%%%%%%%%%%%%%%%%
\begin{figure}[t] % 'r' for right, 'l' for left
    \centering
    \includegraphics[width=\linewidth]{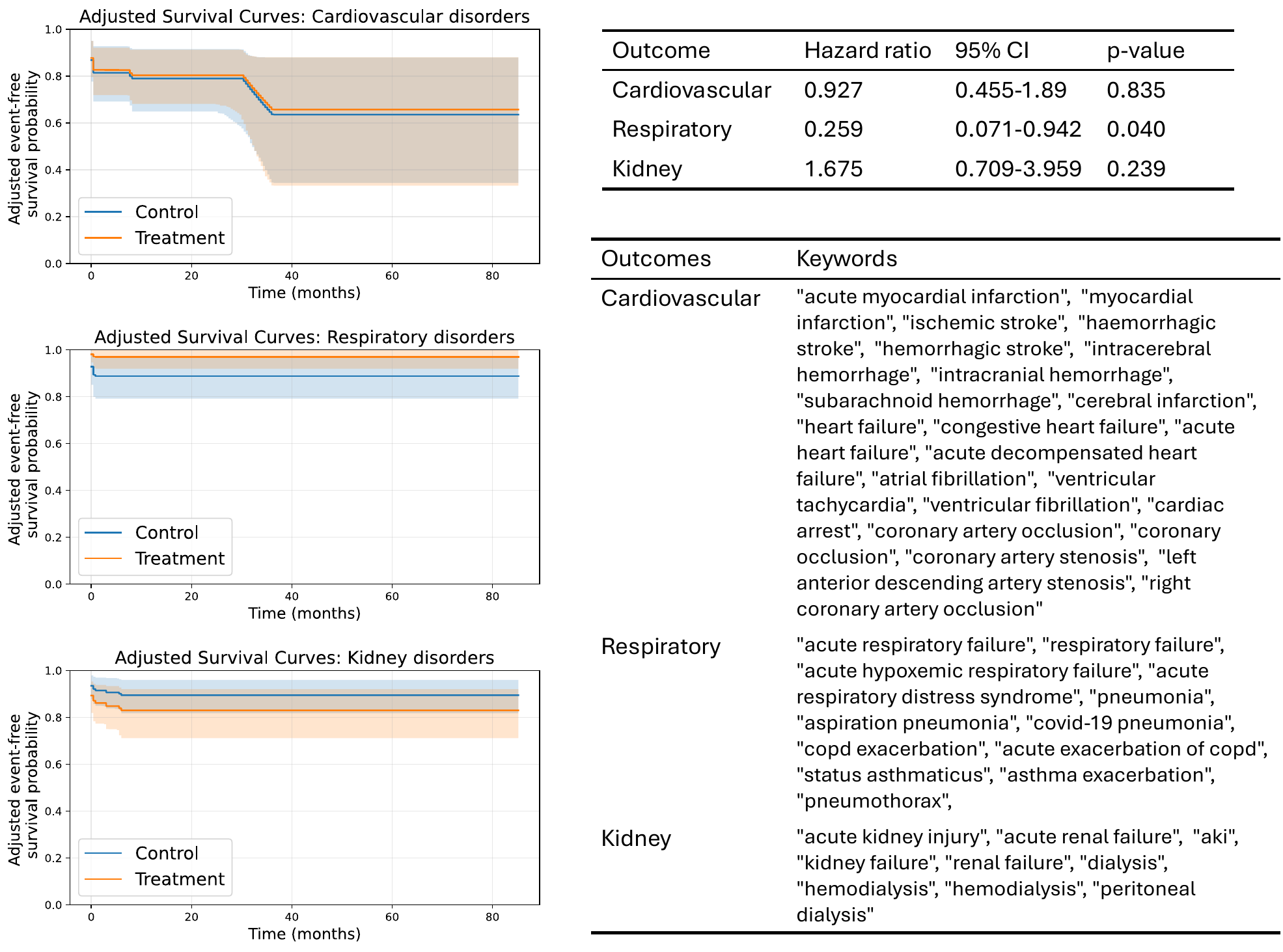} % Adjust width
    \caption{Time-to-onset survival modeling using GLP-1RA textual time series. \textbf{Left}: age/sex-adjusted event-free survival curves from Cox proportional hazards models for cardiovascular, respiratory, and kidney outcomes (treatment/control: diabetes patients with/without GLP medication exposure). Shaded bands denote uncertainty for the adjusted curves. \textbf{Right}: corresponding adjusted hazard ratios (95\% CI, p-value) and the keyword lexicons used to identify outcomes from timeline events.
    }
    \label{fig:survival}
    \vspace{-2mm}
\end{figure}
%%%%%%%%%%%%%%%%%%%%%%%%%%%%%%%%%%%%%%%%%%%%%%%%%%%%%%%%%%%
\vspace{-2mm}
\subsection*{Time-to-onset survival modeling}
\vspace{-2mm}

%In a downstream demonstration of clinical utility, we used the extracted timelines to perform time-to-onset survival modeling for three outcome categories (cardiovascular, respiratory, and kidney).
Figure~\ref{fig:survival} (left) shows covariate-adjusted event-free survival curves from Cox proportional hazards models (adjusted for age and sex), comparing diabetes case reports with early GLP-1RA exposure to a comparison control cohort. Overall, the adjusted curves suggest outcome-specific differences: respiratory outcomes show the largest separation between groups, whereas cardiovascular and kidney outcomes exhibit substantial overlap over follow-up.

Consistent with the curves, the age/sex-adjusted Cox models (Figure~\ref{fig:survival}, right) indicate no clear association between GLP-1RA exposure status and time to first cardiovascular outcome event in the extracted timeline (HR $=0.927$, 95\% CI $0.455$--$1.89$, $p=0.835$). For respiratory outcomes, GLP-1RA exposure is associated with a lower hazard of first documented outcome event (HR $=0.259$, 95\% CI $0.071$--$0.942$, $p=0.040$), which is directionally consistent with prior reports of improved respiratory outcomes among GLP-1RA users \cite{foer2023association}. For kidney outcomes, the point estimate suggests a higher hazard (HR $=1.675$), but uncertainty is large and the association is not statistically significant (95\% CI $0.709$--$3.959$, $p=0.239$). This direction differs from kidney-protective associations commonly reported in larger GLP-1RA studies and may reflect case-report selection effects, limited covariate adjustment, and keyword-based outcome ascertainment from narrative timelines \cite{badve2025effects}. These hazard ratios are interpreted as associative rather than causal, given the observational nature of case reports and incomplete capture of confounders.

%%%%%%%%%%%%%%%%%%%%%%%%%%%%%%%%%%%%%%%%%%%%%%%%%%%%%%%%%%%
%%%%%%%%%%%%%%%%%%%%%%%%%%%%%%%%%%%%%%%%%%%%%%%%%%%%%%%%%%%
\section*{Discussion and Conclusion}
\vspace{-2.5mm}
We present a novel GLP-1RA temporal corpus constructed from PubMed Open Access case reports by converting narratives into clinical textual time series. Using an LLM-as-annotator pipeline, we generate timelines with multiple LLMs and curate a two-expert manual gold standard for evaluation. We will release both the multi-LLM annotations and the expert gold standard timelines as a pilot benchmark corpus upon acceptance. Our evaluation shows that higher-capability reasoning models like GPT5 achieve the strongest overall balance between event matching and temporal fidelity relative to the gold standard. Descriptive analyses indicate that the cohort is dominated by expected cardiometabolic comorbidities (e.g., obesity, hypertension, diabetes) with long-horizon, heterogeneous trajectories. Finally, we demonstrate downstream utility of our clinical timelines by conducting time-to-onset survival analyses for respiratory, cardiovascular, and kidney outcomes, highlighting outcome-specific differences in association with GLP-1RA exposure.

Our study has certain limitations to consider. \textbf{First}, our textual time-series corpus for GLP is a clinical research informatics cohort and should not be considered a representative sample. Corpora derived from case reports are often subject to publication bias and over-representation of rare findings. \textbf{Second}, manual annotations by expert clinicians are costly and time-intensive, limiting the size of the reference set and the breadth of outcomes that can be validated at expert level. Nevertheless, we expand our manual annotation pool from our prior work \cite{wang2025large} (n=10) to having double annotations on 136 case reports. \textbf{Third}, because outcomes are ascertained from keyword-matched timeline events, including baseline mentions at $t{=}0$, estimated time-to-onset may reflect time to first documentation in the narrative rather than biological onset. \textbf{Finally}, our pipeline relies heavily on LLMs for cohort identification and temporal extraction; while this enables scalability, it can introduce subtle extraction and timestamping errors that are difficult to fully characterize and may propagate into downstream analyses.
% Two GLP-specific limitations are important to consider. Fourth, outcome onset is identified using keyword lexicons applied to extracted event strings, which can miss synonymous or indirect mentions and can introduce outcome and timing misclassification. Fifth, our survival models adjust only for age and sex from the narratives; unmeasured confounding and incomplete clinical context mean the estimated hazard ratios should be interpreted as associative, not causal.

Our methodology is not specific to GLP-1RAs and can be extended to other clinical conditions with minimal changes to the core pipeline. We focused on GLP-1RA case reports because they provide a clinically important, medication-centered setting where narratives often describe long-horizon trajectories (e.g., comorbidity evolution, treatment adjustments, and downstream outcomes) that are difficult to represent using structured data alone. This cohort therefore serves as a useful testbed for evaluating whether LLMs can extract clinically meaningful findings and align them temporally over extended time scales. At the same time, the same extraction, matching, and temporal-evaluation framework can be “plugged in” to other disease-focused corpora by changing the cohort-identification criteria and outcome definitions.

Generalizing across conditions raises several practical considerations. First, the relevant time scales differ substantially by clinical domain: chronic or longitudinal conditions may span months to years (as in GLP-1RA case reports), whereas critical-care conditions such as sepsis often evolve over hours to days, yielding narrower time distributions and potentially different failure modes for timestamp assignment. Second, the amount and type of temporal evidence in narratives can vary by report genre; for example, imaging-focused case reports may describe a single snapshot in time with limited longitudinal structure, reducing the value of timeline reconstruction. Finally, as time scales broaden, absolute timestamp errors naturally grow, which motivates discrepancy measures that remain comparable across acute and chronic settings (e.g., log-scaled summaries such as AULTC) rather than being dominated by long-horizon cases.

For temporal modeling, text-derived timelines offer a valuable complement to commonly used structured longitudinal datasets. Continued progress in temporal extraction can further improve this pipeline and can be tuned to different downstream goals, since “high-quality” time series depends on the application—for example, correct event ordering is essential for mechanistic interpretation and causal reasoning, whereas accurate timestamps are especially important for forecasting, alignment, and time-to-event analyses.

Although our event–time pairs aim to approximate when clinical events occurred, another useful framing is the timing of perceived events—when information becomes available to clinicians or patients and begins to influence decisions. This perspective may better capture real-world processes around GLP-1RA care, such as symptom recognition, adherence changes, dose adjustments, and management of adverse effects. Looking ahead, multimodal extensions that combine narrative timelines with complementary signals (e.g., laboratory trends, imaging findings, or other structured measurements when available) could help reconcile discrepancies between documented and perceived time, strengthen early detection of clinically meaningful changes, and support better monitoring and risk stratification strategies.
%%%%%%%%%%%%%%%%%%%%%%%%%%%%%%%%%%%%%%%%%%%%%%%%%%%%%%%%%%%
%%%%%%%%%%%%%%%%%%%%%%%%%%%%%%%%%%%%%%%%%%%%%%%%%%%%%%%%%%%

\subparagraph{Acknowledgments}
This research was supported by the Intramural Research Program of the National Institutes of Health (NIH) and utilized the computational resources of the \href{http://hpc.nih.gov}{NIH HPC Biowulf cluster}. The contributions of the NIH author(s) are considered Works of the United States Government. The findings and conclusions presented in this paper are those of the author(s) and do not necessarily reflect the views of the NIH or the U.S. Department of Health and Human Services. 

% References as numbers
\makeatletter
\renewcommand{\@biblabel}[1]{\hfill #1.}
\makeatother

% unstr is used to keep citation order
\bibliographystyle{vancouver}
\bibliography{references}  

\end{document}